\documentclass[letterpaper, 10 pt, conference]{ieeeconf}  
\usepackage[size=footnotesize]{caption}
\usepackage[noadjust]{cite}
\def\tablename{Table}
\usepackage{amsmath}
\usepackage[export]{adjustbox}
\usepackage{bm}
\usepackage{graphicx}
\usepackage{amssymb}
\usepackage{svg}
\usepackage{array}
\usepackage{url}
\usepackage{hyperref}
\usepackage{lipsum}
\usepackage{tabularx}
\usepackage[capitalise]{cleveref}
\crefformat{equation}{(#2#1#3)}
\usepackage{tikz}
\usetikzlibrary{decorations.pathreplacing,calc}
\newcommand{\tikzmark}[1]{\tikz[overlay,remember picture] \node (#1) {};}

\newcommand*{\AddNote}[4]{%
    \begin{tikzpicture}[overlay, remember picture]
        \draw [decoration={brace,amplitude=0.5em},decorate,ultra thick]
            ($(#3)!(#1.north)!($(#3)-(0,1)$)$) --  
            ($(#3)!(#2.south)!($(#3)-(0,1)$)$)
                node [align=center, text width=1cm, pos=0.5, anchor=west,scale=0.8] {#4};
    \end{tikzpicture}
}%
\usepackage{tikz-3dplot}
\tikzset{small dot/.style={fill=black,circle,outer sep=8pt,scale=0.25}}
\usetikzlibrary{decorations.text}
\usetikzlibrary{shapes.multipart}
\makeatletter

\newif\ifpgfcirclecrosssplitcustomfill
\tikzset{%
	circle cross split part fill/.code=\def\pgf@lib@sh@ccs@list@fill{#1}\pgfcirclecrosssplitcustomfilltrue,%
	circle cross split uses custom fill/.is if=pgfcirclecrosssplitcustomfill}
\pgfdeclareshape{circle cross split}{%
	\nodeparts{text,two,three,four}%
	\savedanchor\centerpoint{%
		\pgfmathsetlength\pgf@xa{\pgfkeysvalueof{/pgf/inner xsep}}%
		\pgfmathsetlength\pgf@ya{\pgfkeysvalueof{/pgf/inner ysep}}%
		\pgf@x\wd\pgfnodeparttextbox
		\pgf@yb\dp\pgfnodeparttextbox
		\pgf@yc\dp\pgfnodeparttwobox
		\ifdim\pgf@yb>\pgf@yc
		\pgf@yc\pgf@yb
		\fi
		\advance\pgf@y-\pgf@yc
		\advance\pgf@x\pgf@xa
		\advance\pgf@y-\pgf@ya
		\advance\pgf@x.5\pgflinewidth
		\advance\pgf@y-.5\pgflinewidth
	}%
	\savedanchor\twoanchor{%
		\pgfmathsetlength\pgf@xa{\pgfkeysvalueof{/pgf/inner xsep}}%
		\pgfmathsetlength\pgf@ya{\pgfkeysvalueof{/pgf/inner ysep}}%
		\advance\pgf@x.5\pgflinewidth
		\advance\pgf@x\pgf@xa
		\advance\pgf@y.5\pgflinewidth
		\advance\pgf@y\pgf@ya
		\pgf@yb\dp\pgfnodeparttextbox
		\pgf@yc\dp\pgfnodeparttwobox
		\ifdim\pgf@yb>\pgf@yc
		\pgf@yc\pgf@yb
		\fi
		\advance\pgf@y\pgf@yc
	}%
	\savedanchor\threeanchor{%
		\pgfmathsetlength\pgf@ya{\pgfkeysvalueof{/pgf/inner ysep}}%
		\pgf@x\wd\pgfnodeparttextbox
		\pgf@yb\dp\pgfnodeparttextbox
		\pgf@yc\dp\pgfnodeparttwobox
		\ifdim\pgf@yb>\pgf@yc
		\pgf@yc\pgf@yb
		\fi
		\advance\pgf@y-\pgf@yc
		\advance\pgf@y-2\pgf@ya
		\advance\pgf@y-\pgflinewidth
		\pgf@yb\ht\pgfnodepartthreebox
		\pgf@yc\ht\pgfnodepartfourbox
		\ifdim\pgf@yb>\pgf@yc
		\pgf@yc\pgf@yb
		\fi
		\advance\pgf@y-\pgf@yc
		\advance\pgf@x-\wd\pgfnodepartthreebox
	}%
	\savedanchor\fouranchor{%
		\pgfmathsetlength\pgf@xa{\pgfkeysvalueof{/pgf/inner xsep}}%
		\advance\pgf@x\wd\pgfnodepartthreebox
		\advance\pgf@x2\pgf@xa
		\advance\pgf@x\pgflinewidth
	}%
	\saveddimen\radius{%
		\pgf@y\ht\pgfnodeparttextbox
		\pgf@yb\ht\pgfnodeparttwobox
		\ifdim\pgf@yb>\pgf@y
		\pgf@y\pgf@yb
		\fi
		\pgf@yc\dp\pgfnodeparttextbox
		\pgf@yb\dp\pgfnodeparttwobox
		\ifdim\pgf@yc>\pgf@yb
		\advance\pgf@y\pgf@yc
		\else
		\advance\pgf@y\pgf@yb
		\fi
		\pgf@yb\ht\pgfnodepartthreebox
		\ifdim\pgf@yb<\ht\pgfnodepartfourbox
		\pgf@yb\ht\pgfnodepartfourbox
		\fi
		\pgf@yc\dp\pgfnodepartthreebox
		\ifdim\pgf@yc<\dp\pgfnodepartfourbox
		\advance\pgf@yb\dp\pgfnodepartfourbox
		\else
		\advance\pgf@yb\pgf@yc
		\fi
		\ifdim\pgf@yc>\pgf@y
		\pgf@y\pgf@yc
		\fi
		\pgfmathsetlength\pgf@ya{\pgfkeysvalueof{/pgf/inner ysep}}%
		\advance\pgf@y2\pgf@ya
		\pgf@x\wd\pgfnodeparttextbox
		\pgf@xa\wd\pgfnodepartthreebox
		\pgf@xb\wd\pgfnodeparttwobox
		\pgf@xc\wd\pgfnodepartfourbox
		\ifdim\pgf@xa>\pgf@x
		\pgf@x\pgf@xa
		\fi
		\ifdim\pgf@xb>\pgf@x
		\pgf@x\pgf@xb
		\fi
		\ifdim\pgf@xc>\pgf@x
		\pgf@x\pgf@xc
		\fi
		\pgfmathsetlength\pgf@xa{\pgfkeysvalueof{/pgf/inner xsep}}%
		\advance\pgf@x2\pgf@xa
		\ifdim\pgf@y>\pgf@x
		\pgf@x\pgf@y
		\fi
		\advance\pgf@x.5\pgflinewidth
		\pgfmathsetlength{\pgf@xb}{\pgfkeysvalueof{/pgf/minimum width}}%
		\pgfmathsetlength{\pgf@yb}{\pgfkeysvalueof{/pgf/minimum height}}%
		\ifdim\pgf@x<.5\pgf@xb
		\pgf@x=.5\pgf@xb
		\fi
		\ifdim\pgf@x<.5\pgf@yb
		\pgf@x=.5\pgf@yb
		\fi
		\pgfmathsetlength{\pgf@xb}{\pgfkeysvalueof{/pgf/outer xsep}}%
		\pgfmathsetlength{\pgf@yb}{\pgfkeysvalueof{/pgf/outer ysep}}%
		\ifdim\pgf@xb<\pgf@yb
		\advance\pgf@x\pgf@yb
		\else
		\advance\pgf@x\pgf@xb
		\fi
	}%
	\inheritanchorborder[from=circle]%
	\inheritanchor[from=circle]{north}%
	\inheritanchor[from=circle]{north west}%
	\inheritanchor[from=circle]{north east}%
	\inheritanchor[from=circle]{center}%
	\inheritanchor[from=circle]{west}%
	\inheritanchor[from=circle]{east}%
	\inheritanchor[from=circle]{mid}%
	\inheritanchor[from=circle]{mid west}%
	\inheritanchor[from=circle]{mid east}%
	\inheritanchor[from=circle]{base}%
	\inheritanchor[from=circle]{base west}%
	\inheritanchor[from=circle]{base east}%
	\inheritanchor[from=circle]{south}%
	\inheritanchor[from=circle]{south west}%
	\inheritanchor[from=circle]{south east}%
	\anchor{two}{\twoanchor}%
	\anchor{three}{\threeanchor}%
	\anchor{four}{\fouranchor}%
	\inheritbackgroundpath[from=circle]
	\beforebackgroundpath{%
		\pgfutil@tempdima=\radius
		\pgfmathsetlength{\pgf@xb}{\pgfkeysvalueof{/pgf/outer xsep}}%
		\pgfmathsetlength{\pgf@yb}{\pgfkeysvalueof{/pgf/outer ysep}}%
		\ifdim\pgf@xb<\pgf@yb
		\advance\pgfutil@tempdima by-\pgf@yb
		\else
		\advance\pgfutil@tempdima by-\pgf@xb
		\fi
		\advance\pgfutil@tempdima by-.5\pgflinewidth%
		\pgfsetshortenstart{0pt}%
		\pgfsetshortenend{0pt}%
		\pgfsetarrows{-}%
		\pgfpathmoveto{\pgfpointadd{\centerpoint}{\pgfqpoint{-\pgfutil@tempdima}{0pt}}}%
		\pgfpathlineto{\pgfpointadd{\centerpoint}{\pgfqpoint{\pgfutil@tempdima}{0pt}}}%
		\pgfpathmoveto{\pgfpointadd{\centerpoint}{\pgfqpoint{0pt}{-\pgfutil@tempdima}}}%
		\pgfpathlineto{\pgfpointadd{\centerpoint}{\pgfqpoint{0pt}{\pgfutil@tempdima}}}%
		\pgfusepathqstroke
	}%
	\behindbackgroundpath{%
		\pgfutil@tempdima=\radius
		\pgfmathsetlength{\pgf@xb}{\pgfkeysvalueof{/pgf/outer xsep}}%
		\pgfmathsetlength{\pgf@yb}{\pgfkeysvalueof{/pgf/outer ysep}}%
		\ifdim\pgf@xb<\pgf@yb
		\advance\pgfutil@tempdima by-\pgf@yb
		\else
		\advance\pgfutil@tempdima by-\pgf@xb
		\fi
		\advance\pgfutil@tempdima by-.5\pgflinewidth%
		\ifpgfcirclecrosssplitcustomfill%
		\pgf@lib@sh@rs@process@list{\pgf@lib@sh@ccs@list@fill}{4}%
		{%
			\pgfmathloop
			\ifnum\pgfmathcounter>4%
			\else%
			\pgf@lib@sh@getalpha\pgf@lib@sh@rs@number{\pgfmathcounter}%
			\edef\pgf@tempa{\csname pgf@lib@sh@rs@\pgf@lib@sh@rs@number @item\endcsname}%
			\ifx\pgf@tempa\pgf@lib@sh@rs@nonetext\else
			\pgfsetfillcolor{\pgf@tempa}%
			\pgf@lib@sh@ccs@angles{\pgfmathcounter}%
			\pgfpathmoveto{\centerpoint}%
			\pgfpathlineto{\pgfpointadd{\centerpoint}{\pgfqpointpolar{\pgf@lib@sh@ccs@angle}{\pgfutil@tempdima}}}%
			\pgfpatharc{\pgf@lib@sh@ccs@angle}{\pgf@lib@sh@ccs@angle@}{\pgfutil@tempdima}%
			\pgfpathclose
			\pgfusepathqfill
			\fi
			\repeatpgfmathloop
		}%
		\fi%
	}%
}
\def\pgf@lib@sh@ccs@angles#1{%
	\ifcase#1\or\def\pgf@lib@sh@ccs@angle{90}%
	\or\def\pgf@lib@sh@ccs@angle{0}%
	\or\def\pgf@lib@sh@ccs@angle{180}%
	\else\def\pgf@lib@sh@ccs@angle{270}%
	\fi
	\edef\pgf@lib@sh@ccs@angle@{\number\numexpr\pgf@lib@sh@ccs@angle+90\relax}%
}
\makeatother
\usepackage{booktabs}

\usepackage[ruled]{algorithm}
\usepackage{algorithmic}

\IEEEoverridecommandlockouts                              

\title{\LARGE \bf
Constraint Handling in Continuous-Time DDP-Based \\ Model Predictive Control
}

\author{Jean-Pierre Sleiman, Farbod Farshidian, Marco Hutter
\thanks{This research was supported in part by the Swiss National Science Foundation through the National Centre of Competence in Research Robotics (NCCR Robotics), and in part by TenneT.}
\thanks{All authors are with the Robotic Systems Lab, ETH Zurich, Zurich 8092, Switzerland. 
(Email: {\tt\small  jsleiman@ethz.ch})
}
}

\begin{document}

\maketitle
\thispagestyle{empty}
\pagestyle{empty}

\begin{abstract}
The Sequential Linear Quadratic (SLQ) algorithm is a continuous-time version of the well-known Differential Dynamic Programming (DDP) technique with a Gauss-Newton Hessian approximation. This family of methods has gained popularity in the robotics community due to its efficiency in solving complex trajectory optimization problems. However, one major drawback of DDP-based formulations is their inability to properly incorporate path constraints. In this paper, we address this issue by devising a constrained SLQ algorithm that handles a mixture of constraints with a previously implemented projection technique and a new augmented-Lagrangian approach. By providing an appropriate multiplier update law, and by solving a single inner and outer loop iteration, we are able to retrieve suboptimal solutions at rates suitable for real-time model-predictive control applications. We particularly focus on the inequality-constrained case, where three augmented-Lagrangian penalty functions are introduced, along with their corresponding multiplier update rules. These are then benchmarked against a relaxed log-barrier formulation in a cart-pole swing up example, an obstacle-avoidance task, and an object-pushing task with a quadrupedal mobile manipulator. 
\end{abstract}
\section{INTRODUCTION} \label{Introduction}
Model Predictive Control (MPC) is a prominent and well-established technique that combines continuous feedback with a lookahead strategy to synthesize stabilizing actions for a broad range of dynamical systems. Its ability to encode complex high-level tasks in simple and intuitive cost functions, while accounting for system constraints, has made it quite compelling in the robotics community. For instance, with regards to locomotion research, this approach has proven its effectiveness in generating dynamic motions for highly articulated underactuated machines such as humanoids \cite{HRP2,Tassa} or quadrupeds \cite{MitCheetah,Neunert,Farbod2,Ruben1,RAL}.
Fundamentally, MPC operates by repeatedly solving a finite-horizon optimal control problem (OCP) in a receding-horizon fashion. It is therefore clear that the quality of the resulting control law heavily relies on the underlying optimal control formulation and on the scheme used to solve it \cite{Betts}. These two components dictate how much of the problem's true complexity is captured in the formulation, in addition to the speed at which optimal trajectories are calculated. Direct Trajectory Optimization (TO) approaches transcribe the OCP through a time-discretization of the states and inputs; thereby transforming the infinite-dimensional optimization problem into a finite-dimensional one that could be solved with standard nonlinear programming (NLP) solvers. These optimization-based methods have drawn great interest due to their ability to naturally incorporate any form of path constraints. However, they typically carry a high computational burden that renders them inapplicable in real-time settings, with a few notable exceptions that tend to exploit the problem's sparse structure \cite{Boyd, Domahidi}. 

In contrast, indirect methods rely on fundamental principles that provide necessary or sufficient conditions of optimality to solve the original optimal control problem. Particularly, one such method that has gained significant traction recently is Differential Dynamic Programming \cite{Mayne}. This technique relies on \emph{Bellman's principle of optimality} \cite{Bellman} to decompose the problem into smaller minimization sub-problems that are solved recursively. To avoid the ``curse-of-dimensionality'' attributed to Dynamic Programming, DDP uses local quadratic approximations of the stage cost, dynamics, and cost-to-go, around nominal state-input trajectories to compute an affine control sequence from a backward Riccati equation. Variants of the DDP method such as the iterative Linear Quadratic Regulator (iLQR) \cite{ilqr1,ilqr2} or the Sequential Linear Quadratic algorithm \cite{Sideris,Farbod1} follow a similar mechanism but use first-order approximations of the dynamics instead, which reduces the computation time at the expense of slower convergence rates. This family of TO schemes has a linear complexity with respect to the time horizon, which makes it favorable in real-time control applications \cite{Tassa2,Farbod2,Neunert2}. However, unlike NLP solvers, the Riccati solvers used by DDP-based methods are not inherently designed to handle constraints. 

Most researchers tackling the constrained-DDP problem have been inspired by concepts from the well-developed literature on optimization theory \cite{Bertsekas,Nocedal}. For instance, Tassa et al. \cite{ControlLimited} accomodate box-constraints on the inputs by using a projected-Newton method to successively solve a sequence of small quadratic programs (QP) in the backward pass. In \cite{Zhaoming}, the optimal inputs are ensured to be constraint-consistent by solving a Karush-Kuhn-Tucker (KKT) system that contains the active set. A QP is then solved in the forward pass to guarantee that the updated nominal trajectories, under the affine control law, are still feasible. Other works have adopted an augmented-Lagrangian (AL) approach \cite{Bertsekas} to transform the generic constrained-DDP problem into an unconstrained one \cite{ALTRO,Plancher}. Certain authors have attempted to combine different optimization-based notions in a single framework: Lantoine et al. \cite{Lantoine} use an active-set method along with an augmented-Lagrangian formulation to handle hard and soft inequality constraints, respectively; while in \cite{Aoyama}, the solver switches between AL-DDP and a primal-dual interior point method to exploit the benefits carried by both approaches. 

We specifically focus on the continuous-time, constrained SLQ algorithm introduced in \cite{Farbod1}. The method relies on a projection technique and a penalty function to handle state-input and state-only equality constraints, respectively. Moreover, it has later been extended to incorporate inequality constraints via a relaxed log-barrier function \cite{Ruben1}. In this paper, we propose an augmented-Lagrangian SLQ variant
(AL-SLQ) to overcome the numerical ill-conditioning issues typically associated with penalty and barrier methods \cite{Bertsekas,Nocedal}. We also provide important considerations regarding the AL inner/outer loop updates that allow us to run our algorithm within a real-time MPC scheme. Finally, a comparative study is presented among different inequality-handling methods both in the theoretical and experimental sections of this paper. The experimental part includes results from numerical simulations of various constrained robotic tasks.    

\section{BACKGROUND AND MOTIVATION} \label{sec:Background}
In this section, we provide the reader with preliminary ideas that support and motivate the development of the constrained SLQ algorithm in~\cref{sec:ConstrainedSLQ}.
\subsection{Unconstrained SLQ} \label{sec:SLQ}
We consider the representation of a generic dynamical system through a set of nonlinear differential equations 
\begin{equation} \label{eq:ode}
    \dot{\bm x}(t) = \bm f (\bm x(t), \bm u(t), t)
\end{equation}
with $\bm x \in \mathbb{R}^{n_x}$ denoting the state variables, $\bm u \in \mathbb{R}^{n_u}$
the input variables, and $\bm f(\bm x, \bm u, t) \in \mathbb{R}^{n_x}$ a continuously differentiable flow map. An unconstrained OCP for such systems can be devised as follows
\begin{equation} \label{eq:OCP}
    \begin{cases}
    \underset{\bm u(.)}{\min} \ \ \Phi(\bm x(t_f)) + \displaystyle \int_{t_0}^{t_f} L(\bm x(t), \bm u(t), t)dt \\[2ex]
    \text{s.t.} \ \ \dot{\bm{x}}(t) = \bm f(\bm x(t), \bm u(t), t) \ \ \ \ \forall t\in[t_0,t_f] \\
    \ \ \ \ \ \bm x(t_0) = \bm x_0
    \end{cases}
\end{equation} 
where the functional being minimized consists of an intermediate cost ${L(\bm x, \bm u, t)}$ and a terminal cost ${\Phi(\bm x(t_f))}$.  

SLQ finds optimal trajectories to~\cref{eq:OCP} by iteratively solving time-varying local approximations of the original problem. This requires the computation of first-order derivatives for the dynamics, and quadratic approximations of the objective function around nominal trajectories $\bar{\bm x}(t)$ and $\bar{\bm u}(t)$:
\begin{subequations}
\begin{align} \label{eq:approximations}
    \delta \dot{\bm x} &\approx \bm A(t) \delta{\bm x} + \bm B(t) \delta{\bm u} \\
    \Tilde{\Phi} &\approx \dfrac{1}{2} \delta \bm x^T \bm Q_f \delta \bm x + \bm q_f^T \delta \bm x + q_f \\
    \begin{split}
    \Tilde{L} &\approx \dfrac{1}{2} \delta \bm x^T \bm Q(t) \delta \bm x + \dfrac{1}{2} \delta \bm u^T \bm R(t) \delta \bm u \\
    &+ \delta \bm u^T \bm P(t) \delta \bm x + \bm q^T(t) \delta \bm x + \bm r^T(t) \delta \bm u + q(t)
    \end{split}
\end{align}
\end{subequations} 
where $\delta \bm x$ and $\delta \bm u$ are perturbations from the nominal trajectories. In order to solve the approximate OCP at the current iteration, SLQ relies on \textit{Pontryagin's minimum principle} which provides the necessary conditions for optimality \cite{Stengel}. These entail a two-point boundary value problem (BVP) governed by~\cref{eq:approximations} and $n_x$ differential equations defined with respect to the costate (adjoint) variables $\bm \lambda(t)$ as follows: 
\begin{align} \label{eq:Pontryagin}
    \dot{\bm \lambda}^* = -\nabla_{\delta \bm x} \tilde H,
    \qquad \text{with }
    \bm \lambda^*(t_f) = \nabla_{\delta \bm x} \tilde \Phi\big|_{t = t_f}
\end{align}
$\tilde H$ corresponds to the Hamiltonian of the linear-quadratic OCP and is defined as 
\begin{equation}
   \tilde H(\delta \bm x, \delta \bm u, \bm \lambda, t) :
   = \tilde L + \bm \lambda^T \left(\bm A \delta{\bm x} + \bm B \delta{\bm u}\right)
\end{equation}
The optimal input variables are retrieved by minimizing the Hamiltonian function 
\begin{equation}\label{eq:hamiltonian}
    \delta \bm u^* = \underset{\delta \bm u}{\text{arg\,min }} \tilde H 
\end{equation}
Furthermore, by ensuring that ${\nabla^2_{\delta \bm u}\tilde H = \bm R(t)}$ is always positive definite, we obtain sufficient conditions of optimality from the strengthened \textit{Legendre-Clebsch convexity condition} \cite{Stengel}. This allows us to derive a closed-form solution for~\cref{eq:hamiltonian} from the expression below
\begin{equation}
    \nabla_{\delta \bm u} \tilde H \big |_{\delta \bm u = \delta \bm u^*} = 0
\end{equation}
By using a proper assumption on the form of the costate variables ${\bm \lambda(t) = \bm S(t) \delta \bm x + \bm s(t)}$, and then mathematically manipulating the BVP, a differential Riccati equation emerges \cite{Farbod1}. Integrating the Riccati equation backwards in time results in an affine control policy
\begin{equation} \label{eq:Policy}
    \delta \bm u^*(t) = \delta \bm u_{ff}(t) + \bm K(t) \delta \bm x(t)
\end{equation}
which is applied to~\cref{eq:ode} to generate the new nominal trajectories needed for the next iteration. It is worth noting that by keeping the above derivations in the continuous-time domain, the forward and backward passes can be performed with an adaptive step-size simulator, which provides proper control over the integrator's local truncation error.  

\subsection{Constrained Nonlinear Optimization}
We now suppose we are given a constrained minimization problem with respect to a finite-dimensional vector $\bm z \in \mathbb{R}^{n_z}$
\begin{equation} \label{eq:NLP}
    \begin{cases}
    \underset{\bm z}{\min} \ \ f(\bm z) \\[1ex]
    \text{s.t.} \ \ g_i(\bm z) = 0 \ \ \ \ \ \ \forall i=1,...,n_{eq}\\
    \ \ \ \ \ h_i(\bm z) \geq 0 \ \ \ \ \ \ \forall i=1,...,n_{ineq}
    \end{cases}
\end{equation} 
A popular technique to handle the constraints in~\cref{eq:NLP} is based on transforming the above problem into its equivalent unconstrained version and solving that instead. The equivalence is maintained by absorbing the constraints into the cost function through the use of indicator functions, as follows
\begin{equation} \label{eq:UNLP}
    \underset{\bm z}{\min} \ \ f(\bm z) + \sum_{i=1}^{n_{eq}} \mathcal{I}_{G_i}(\bm z) + \sum_{i=1}^{n_{ineq}} \mathcal{I}_{H_i}(\bm z)
\end{equation}
where\\
\hspace*{-0.2375cm}
\begin{tabularx}{0.5125\textwidth}{XX}
{\begin{align*}
\mathcal{I}_{G_i} := 
&\begin{cases}
0 \quad &\text{if } g_i = 0 \\
+\infty \quad &\text{otherwise}
\end{cases}
\end{align*}} 
& 
{\begin{align*}
\mathcal{I}_{H_i} :=
&\begin{cases}
0 \quad &\text{if } h_i \geq 0 \\
+\infty \quad &\text{otherwise}
\end{cases}
\end{align*}} 
\end{tabularx}
In this discussion, we specifically consider second-order Newton-type algorithms for solving the resulting unconstrained problem. Such methods assume the existence of gradient and Hessian information in the objective. To that end, the indicator functions in~\cref{eq:UNLP} are replaced with differentiable approximations such as the quadratic penalty ${\mathcal{Q}_i = \dfrac{\rho}{2} \cdot ||g_i||^2}$ for equality constraints and the log-barrier function ${\mathcal{B}_i = -\mu \ln(h_i)}$ for inequalities. The parameters ${\rho > 0}$ and ${\mu > 0}$ are weighting factors that are monotonically adapted throughout the optimization iterations. This effectively leads to successive minimizations of simpler perturbed versions of~\cref{eq:UNLP} that ultimately tend to the original problem as ${\rho \rightarrow +\infty}$ and ${\mu \rightarrow 0}$. One could easily show that the indicator functions are indeed recovered at the limit: ${\mathcal{Q}_i \rightarrow \mathcal{I}_{G_i}}$ and ${ \mathcal{B}_i \rightarrow \mathcal{I}_{H_i}}$. 

\subsubsection{Numerical Issues} \label{subsubsec:IllConditioning}
Penalty and barrier methods are typically studied separately in the optimization literature, due to the different ways in which their iterates evolve and converge to the optimal solution \cite{Nocedal}. However, they both suffer from similar issues of ill-conditioning that arise as the relaxed problem approaches the true one, thus causing convergence difficulties. This can be seen in the Hessian expression which -- when $\bm z$ is in the vicinity of the minimizer (for large $\rho$ and small $\mu$) -- can be approximated as
\begin{equation} \label{eq:Hessian}
    \bm H \approx \nabla^2\mathcal{L} + \sum_{i=1}^{n_{eq}} \rho \nabla g_i \nabla g_i^T + \sum_{i=1}^{n_{ineq}} \dfrac{\mu}{h_i^2} \nabla h_i \nabla h_i^T,
\end{equation}
where the first term is the Hessian of the Lagrangian $\mathcal{L}$ for problem~\cref{eq:NLP}. The remaining two terms are responsible for the aforementioned ill-conditioning: Assuming that at the solution $m$-inequality constraints are active, such that ${0 < m + n_{eq} < n_z}$, then $m + n_{eq}$ eigenvalues are very large while the remaining ones are zero. This results in a high condition number for the matrix $\bm H$. 

It is important to note that other approximating functions can be used to encode constraint satisfaction in the cost function. For instance, one that is particularly interesting for the results section of this paper, is the relaxed log-barrier 
\begin{equation} \label{eq:RelaxedBarrier}
    \mathcal{\hat B}(h_i) =
    \begin{cases}
    -\mu \ln(h_i) \quad  h_i > \delta \\
    \ \ \beta(h_i; \delta) \quad h_i\leq \delta
    \end{cases}
\end{equation}
where $\delta$ is a relaxation parameter that separates between a log-barrier and an external penalty function $\beta(h_i;\delta)$. The latter could be a quadratic \cite{Ruben1}, general polynomial or exponential function \cite{RelaxedLogBarrier} designed such that $\mathcal{C}^2$-continuity of $\mathcal{\hat B}$ is maintained at $\delta$. The relaxed barrier is defined for all values of $h_i$, thereby getting rid of the singularity at the boundary of the feasible region, and allowing for infeasible iterates to take place without any algorithmic failures. In order to ensure that for any fixed barrier parameter $\mu$ the solution to the relaxed problem doesn't violate the constraints, $\delta$ has to be smaller than a certain threshold. However, similar to our previous reasoning, the optimal solution to the original problem can only be retrieved as ${\mu\rightarrow 0}$ and ${\delta\rightarrow 0}$, both of which could lead to the same Hessian ill-conditioning issues perceived in~\cref{eq:Hessian}.

\subsubsection{Augmented-Lagrangian}
A powerful remedy to the problems presented in~\cref{subsubsec:IllConditioning} can be found in the augmented-Lagrangian approach. The fundamental idea behind it is to augment the cost with a penalty function ${\mathcal{P}(c(\bm z), \nu, \rho)}$, where $c(\bm z)$ is a generic constraint, $\rho$ is a penalty parameter and $\nu$ is an auxiliary variable that is meant to estimate the optimal Lagrange multiplier corresponding to the constraint $c(\bm z)$. The algorithmic structure consists of an inner loop that uses an unconstrained optimization solver to minimize the augmented-Lagrangian for fixed values of $\rho$ and $\nu$. This is followed by an outer loop that monotonically increases the penalty parameter, and adapts the multiplier estimate with an appropriate predefined update rule
\begin{equation}
    \nu^*_{k+1} = \Pi(c(\bm z^*_{k+1}), \nu_{k}, \rho_k)
\end{equation}
It can be shown that as $\nu$ approaches the true optimal multiplier, the algorithm converges to a Karush-Kuhn-Tucker (KKT) point $(\bm z^*, \nu^*)$ of the original constrained program, which is also a potential primal-dual optimum. In fact, this is true for finite values of the penalty parameter ${\text{(i.e. for any } \rho > \rho_{min})}$ \cite{Bertsekas,Nocedal}; thus implying that convergence can be attained without any numerical issues, through a proper update of the multiplier estimates. We go into further details regarding the chosen penalty functions $\mathcal{P}(.)$ and update rules $\Pi(.)$ in the upcoming section.

\section{CONSTRAINED SLQ-MPC} \label{sec:ConstrainedSLQ}
In this section, we present the main developments behind our constrained SLQ-MPC algorithm for solving general constrained optimal control problems. To begin with, we make use of the notion of \textit{partial elimination of constraints} \cite{Bertsekas}, which allows us to handle each type of constraint differently. To elaborate, inequalities and state-only equalities are treated with an augmented-Lagrangian approach. On the other hand, the dynamic constraint given by~\cref{eq:ode} and state-input equality constraints are adjoined to the cost function through Lagrange multipliers. The former is implicitly satisfied in the SLQ forward rollout, while the algebraic equalities are respected through a projection technique to ensure strict feasibility \cite{Farbod1}. Therefore, the exact same formulation is used here as in \cite{Farbod1}, but with a different cost-functional that includes a constraint-dependent AL-penalty ${\mathcal{P(.)}}$. For the sake of brevity, our subsequent derivations will only cover inequality constraints ${\big(\text{i.e., } \bm h(\bm x, \bm u, t) \geq 0 \big)}$; however, they can readily be extended to the case of pure-state equalities.

\subsection{Inner Loop}
The receding horizon OCP to be solved has the following form:
\begin{equation} \label{eq:ConstrainedOCP}
    \begin{cases}
    \underset{\bm u(.)}{\min} \ \ \underbrace{\Phi(\bm x(t_f)) + \displaystyle \int_{t_0}^{t_f} \big(L(\bm x, \bm u, t) + \mathcal{P}(\bm h, \bm \nu_k, \rho_k) \big) dt}_{\mathcal{L}_A} \\[2ex]
    \text{s.t.} \quad \begin{array}[t]{ll} \dot{\bm{x}} = \bm f(\bm x, \bm u, t) & \forall t\in[t_0,t_f] \\
    \bm g(\bm x, \bm u, t) = 0 & \forall t\in[t_0,t_f] \\
   \bm x(t_0) = \bm x_0 &
    \end{array}
    \end{cases}
\end{equation}
where the initial state $\bm x_0$ is updated at every MPC iteration with the current measured state.
When it comes to selecting a proper AL-penalty, a wide variety of alternatives exist in the literature. Here we focus on three, highlight some of their differences, and compare them in the experimental section. The reader is also referred to the supplementary \href{https://www.youtube.com/watch?v=TXKNEaFvLsk}{\textcolor{blue}{video}}, which supports the explanation presented in this section.

The most popular choice for $\mathcal{P}(.)$ is given by the Powell-Hestenes-Rockafellar (PHR) method \cite{Hestenes,Nocedal}: 
\begin{equation} \label{eq:P1}
    \mathcal{P}_1(\bm h, \bm \nu, \rho) = \sum_{i=1}^{n_{ineq}}\dfrac{1}{2\rho}\big(\max\{0, \nu_i - \rho h_i\}^2 - \nu_i^2 \big)
\end{equation}
This expression is obtained by minimizing the quadratic function ${\mathcal{P} = -\bm \nu^T(\bm h - \bm s) + \dfrac{\rho}{2} ||\bm h - \bm s||^2}$ with respect to ${\bm s \geq 0}$, which are slack variables introduced to transform the inequalities into equivalent equality constraints $\bm h - \bm s = 0$. A different AL-formulation, which we refer to as the non-slack penalty, has been presented in \cite{Toussaint}, and applied in \cite{Plancher} within a DDP context:
\begin{equation} \label{eq:P2}
    \mathcal{P}_2(\bm h, \bm \nu, \rho) = \sum_{i=1}^{n_{ineq}}-\nu_i h_i + \big[h_i < 0 \vee \nu_i > 0\big] \dfrac{\rho}{2} h_i^2 
\end{equation}
It is clear that the terms in equations~\cref{eq:P1} and~\cref{eq:P2} match whenever the corresponding constraint is violated. However, the latter keeps applying the same quadratic penalty in the interior of the feasible region as long as the multiplier estimate is non-zero, regardless of how far the current feasible solution is from the boundary ${(\text{i.e., }h^*_{i_k} \gg 0)}$. Consequently, in the non-slack approach, inequalities are treated as equalities up until the update rule for $\nu_i$ drives it to zero.      

One of the drawbacks of quadratic-type penalties applied to inequality constraints is that the resulting augmented-Lagrangian is not twice differentiable, which might be problematic for second-order methods that are not robust against Hessian discontinuities. Hence, we consider a third option from a family of smooth AL-penalty functions that satisfy certain properties \cite{Bertsekas,ALComparison}:
\begin{equation} \label{eq:P3}
    \mathcal{P}_3(\bm h, \bm \nu, \rho) = \sum_{i=1}^{n_{ineq}} \dfrac{\nu_i^2}{\rho} \psi\bigg(\dfrac{\rho h_i}{\nu_i}\bigg)
\end{equation}
where $\psi(.)$ in our case is defined as a shifted quadratically-relaxed log-barrier \cite{ALComparison}. This somewhat resembles the function~\cref{eq:RelaxedBarrier} used in \cite{Ruben1}, but the advantage in using an AL method with~\cref{eq:P3} is that when a constraint is violated, the corresponding multiplier updates would increase the linear term of the exterior quadratic penalty without affecting the second-order coefficient (which is what typically leads to the numerical issues discussed previously in non-AL methods).
\begin{figure}[!t]
\scalebox{0.9}{
\begin{minipage}{0.5\textwidth}
\begin{algorithm}[H]
\caption{Augmented-Lagrangian SLQ-MPC}
\begin{algorithmic}[1] \label{fig:algorithm}
\STATE Initialize $\bar \rho$, $\alpha$, $\bm u_0(t)$ and $\bm \nu_0(t)$ $\quad \forall t\in [t_0,t_f]$ 
\FOR{$k = 0,1,2,\dots$}
        \STATE \textbf{Forward rollout} of system dynamics $\rightarrow \bm x_k(t)$ \tikzmark{t1}
        \STATE \textbf{Quadratize} augmented-Lagrangian $\mathcal{L}_A(\bm x_k, \bm u_k, \bm \nu_k)$ \tikzmark{r2}
        \STATE \textbf{Linearize} dynamics and state-input equalities 
        \STATE Solve resulting \textbf{Riccati equations}
        \STATE Compute \textbf{control policy}
        \STATE${\quad \rightarrow \bm u_{k+1}(t) = \gamma \cdot \bm u_{ff}(t) + \bm K(t)\delta \bm x(t)}$
        \STATE Perform \textbf{line search} over $\mathcal{L}_A(\bm x_{k+1}, \bm u_{k+1}, \bm \nu_k)$
        \STATE $\quad \rightarrow \gamma^*, \bm u_{k+1}(t; \gamma^*)$  \tikzmark{b1}
        \STATE Update inequality \textbf{multiplier estimates} \tikzmark{t2}
        \STATE $\quad \bm \nu_{k+1} \leftarrow$ $\bm \Pi(\bm h_{k+1}, \bm \nu_k, \bar \rho, \alpha)$
        \STATE Shift $\bm u_{k+1}(t), \bm \nu_{k+1}(t)$ and extrapolate tails\tikzmark{b2}
\ENDFOR
\end{algorithmic}
\AddNote{t1}{b1}{r2}{\small Inner\\Loop}
\AddNote{t2}{b2}{r2}{\small Outer\\Loop}
\end{algorithm}
\end{minipage}}
\vspace{-5mm}
\end{figure}

We recall from~\cref{sec:SLQ} that SLQ iteratively minimizes a linear-quadratic approximation of the OCP. In the case of~\cref{eq:ConstrainedOCP}, this entails a second-order approximation of the augmented-Lagrangian $\mathcal{L}_A$ and thus the inequalities as well. Consequently, both the feedforward and feedback terms of the control policy~\cref{eq:Policy} are shaped by the curvature of these constraints and the current multiplier estimates. Moreover, an Armijo-Goldstein backtracking line-search scheme is applied on the feedforward input-step to further minimize the augmented-Lagrangian function. Finally, it is important to recall that AL-methods rely on a proper minimization of the successive sub-problems (i.e., to ensure optimality up to a prescribed tolerance) each of which is followed by updates in the outer loop. Such a strategy would not be desirable when dealing with real-time applications that require fast and continuous online re-planning. This has been verified before in MPC-related settings, where it was shown that computing suboptimal solutions to each optimization problem with a few iterations, results in superior closed-loop performance when compared to solving them up to optimality at lower rates \cite{Boyd,Diehl}. Similarly, we adopt a real-time iteration scheme \cite{Diehl} to run the constrained SLQ algorithm in a receding-horizon fashion. More specifically, one MPC call consists of a single forward-backward pass in the inner loop (i.e., one Riccati iteration), to update the primal trajectories ${\bm x^*_{k}(t)}$ and ${\bm u^*_{k}(t)}$, followed by a single update of the dual trajectories $\bm \nu^*_{k}(t)$ performed in the outer loop. We only require multiple SLQ calls $(\approx 10)$ during the first MPC iteration to avoid convergence towards bad local minima caused by poor initialization. The constrained SLQ-MPC main loop is summarized in~\cref{fig:algorithm}.         
\subsection{Outer Loop}
As discussed previously, after the inner loop minimization terminates, the penalty parameter $\rho_k$ and the multiplier estimates $\bm \nu_k$ are adapted according to a prespecified update rule. Typically, for standard optimization problems or in open-loop trajectory optimization settings \cite{Plancher}, $\rho$ is updated monotonically as ${\rho_{k+1} = \beta \rho_k}$ where ${\beta > 1}$. In contrast, in the MPC case, penalty parameters (or barrier parameters) \cite{Boyd,Ruben1} are kept constant due to the lack of proper and systematic warm-starting techniques for such algorithms. Similarly, we set ${\rho_k = \bar{\rho}}$ which for AL-based methods would still allow for convergence to the true optimum if $\bar{\rho}$ is large enough.
On the other hand, the update law for the multipliers turns out to be of crucial importance to the convergence of our algorithm. The most common rule is given by ${ \nu_{i_{k+1}} \leftarrow \nabla_{h_i}\mathcal{P} \big |_{\nu_{i_k}, h_{i_{k+1}}} }$ \cite{Nocedal}. This usually involves update steps that are proportional to the penalty parameter. Since in our case we do not follow the standard AL algorithmic procedure, we found it useful in practice to be able to take smaller step-sizes. The reasoning behind this stems from the idea that multiplier updates could be interpreted as gradient ascent steps that are meant to maximize the Lagrange dual function, defined as follows
\begin{equation}
    d(\bm \nu(t)) := \underset{\bm u(.)}{\min} \ \mathcal{L}
\end{equation}
where $\mathcal{L}$ is the Lagrangian of problem~\cref{eq:ConstrainedOCP}. Therefore, the gradient-ascent algorithm is given as
\begin{equation}
    \bm \nu_{k+1} = \bm \nu_k + \alpha \cdot \nabla_{\bm \nu} d
\end{equation}
The parameter $\alpha$ determines the ascent step-length, while the gradient ${\nabla_{\bm \nu} d}$ can be approximated under certain mild assumptions as in \cite{Bertsekas}. For instance, for the PHR penalty function, we get the following update rule -- $\Pi_1$:
\begin{equation} \label{eq:update1}
\begin{split}
    \nu_{i_{k+1}} = \max \big\{ \nu_{i_k} - \alpha h_{i_{k+1}}, (1 - \frac{\alpha}{\rho}) \cdot \nu_{i_k} \big\}
\end{split}
\end{equation}
Two things can be noticed from~\cref{eq:update1}: If we set ${\alpha = \rho}$, then the update rule coincides with the standard one given by ${\nabla_{h_i}\mathcal{P}}$. Moreover, to ensure that ${\nu_{i_k} \rightarrow 0}$ when the corresponding constraint is inactive at the solution, $\alpha$ must satisfy a condition ${(0<\alpha<2\rho)}$ that renders the system ${\nu_{k+1} = (1-\alpha/\rho)\cdot\nu_k}$ asymptotically stable. A similar reasoning is adopted when choosing the update rules for the non-slack penalty~\cref{eq:P2} -- $\Pi_2$:
\begin{equation}
    \nu_{i_{k+1}} = \max\{0,\nu_{i_k} - \alpha h_{i_{k+1}}\}
\end{equation}
and the smooth-PHR penalty~\cref{eq:P3} -- $\Pi_3$:
\begin{equation}
    \nu_{i_{k+1}} = -\alpha \cdot \nu_{i_k} \psi'\bigg(\dfrac{\rho h_{i_{k+1}}}{\nu_{i_k}}\bigg)
\end{equation}
Finally, we note that there exist other update rules given by second-order multiplier methods, or methods that rely on estimating the optimal step-length $\alpha^*$ \cite{Bertsekas}; but those are typically more computationally demanding.  

\section{RESULTS}
In this section, we validate our augmented-Lagrangian SLQ-MPC approach through numerical simulations performed on three different underactuated dynamical systems, with a variety of inequality constraints. We also present comparative results for the AL-penalties introduced in this paper, and benchmark them against the quadratically-relaxed barrier method previously used in \cite{Ruben1}. Supporting results and plots are also included in the accompanying~\href{https://www.youtube.com/watch?v=TXKNEaFvLsk}{\textcolor{blue}{video}} submission. All examples are implemented in C++.  

\subsection{Cart-Pole}
\begin{figure}[t]
\centering
\includegraphics[scale=0.5]{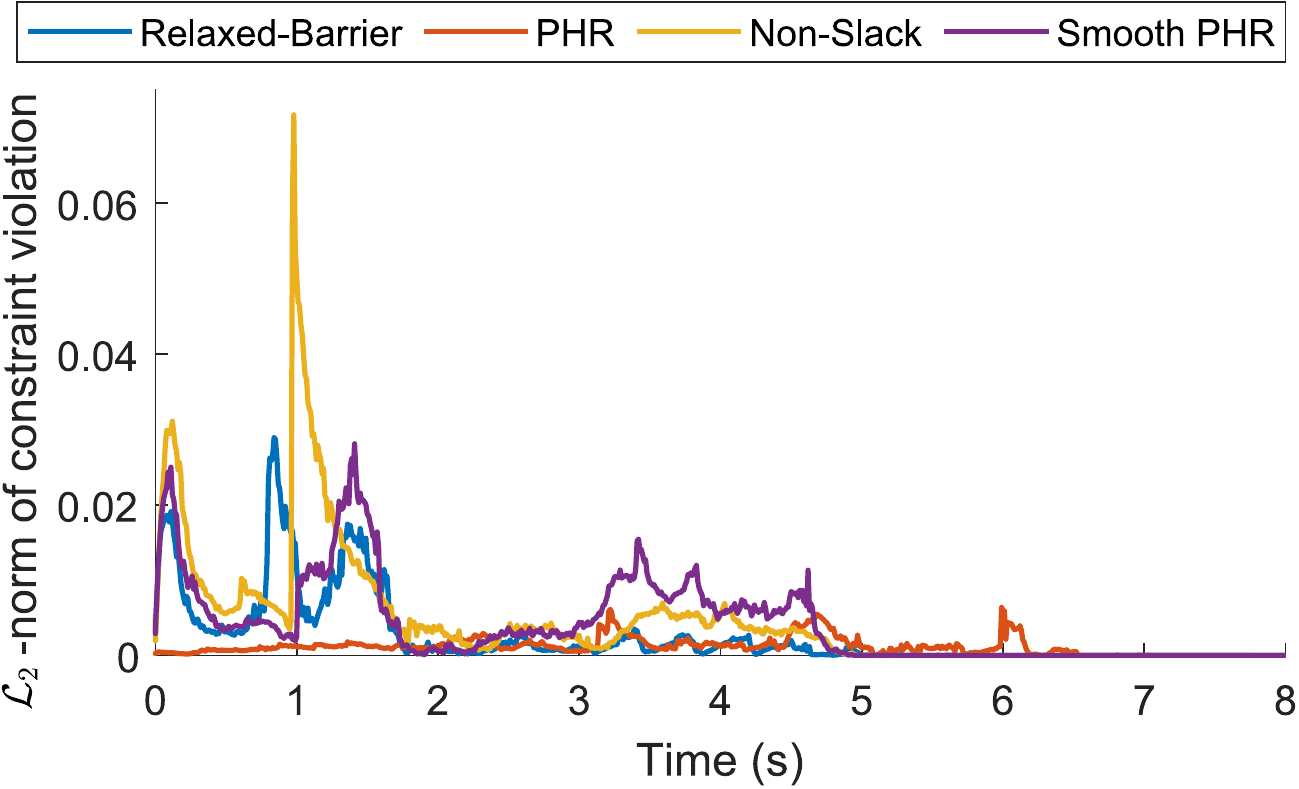}
\caption{Plots showing the $\mathcal{L}_2$-norm of constraint violations for the four methods in the cart-pole example.}
\label{fig:CartpoleConstraints}
\vspace{-7mm}
\end{figure}

We first consider the standard cart-pole swing-up task. The same quadratic cost weights and time horizon ${T = 3 \ \text{s}}$ are used throughout the different test cases. The MPC loop runs at 100 Hz. We impose simple box-constraints ${-u_{max} \leq u \leq u_{max}}$ on the input force acting on the cart. Without any constraints, the force needed to swing the pendulum to its upright position reaches a maximum of ${u \approx 15 \ \text{N}}$ (see input trajectory in \href{https://www.youtube.com/watch?v=TXKNEaFvLsk}{\textcolor{blue}{video}}); therefore, we set a strict limit by choosing ${u_{max} = 5 \ \text{N}}$. The plots in~\cref{fig:CartpoleConstraints} show the $\mathcal{L}_2$-norm of the constraint violations over the time horizon, per MPC iteration, for the four inequality-handling methods. The corresponding parameters are manually tuned in such a way that the best performance (in terms of cost minimization and constraint satisfaction) is attained for each method. We note the following: The relaxed log-barrier works well in this case as it is made stiff by choosing low values for both $\mu$ and $\delta$. This was possible since the number of active constraints was always equal to the number of inputs which, according to our previous discussion, does not lead to any ill-conditioning issues. Moreover, all three AL-penalties manage to realize the swing-up task with minor real-time violations (in the order of ${0.01 \ \text{N}}$), but this was only possible for small step sizes in the multiplier updates. The non-slack penalty led to the poorest performance -- in terms of feasibility and time required for swing-up -- among all four methods. We hypothesize that this is because this penalty~\cref{eq:P2} would still result in gradients that could push feasible iterates towards the boundary of the feasible region, unlike the other penalty functions which tend to vanish when the current iterate satisfies the constraints. Finally, we notice that on average, all four methods have similar computational times; however, the relaxed-barrier and smooth PHR methods are only slightly faster than their counterparts $(\approx2 \ \text{ms faster})$ due to their second-order continuity.    
\subsection{Ballbot}
\begin{figure}[t]
\centering
\begin{minipage}[t]{0.1875\textwidth}
  \centering
    \includegraphics[valign=b,scale=0.096]{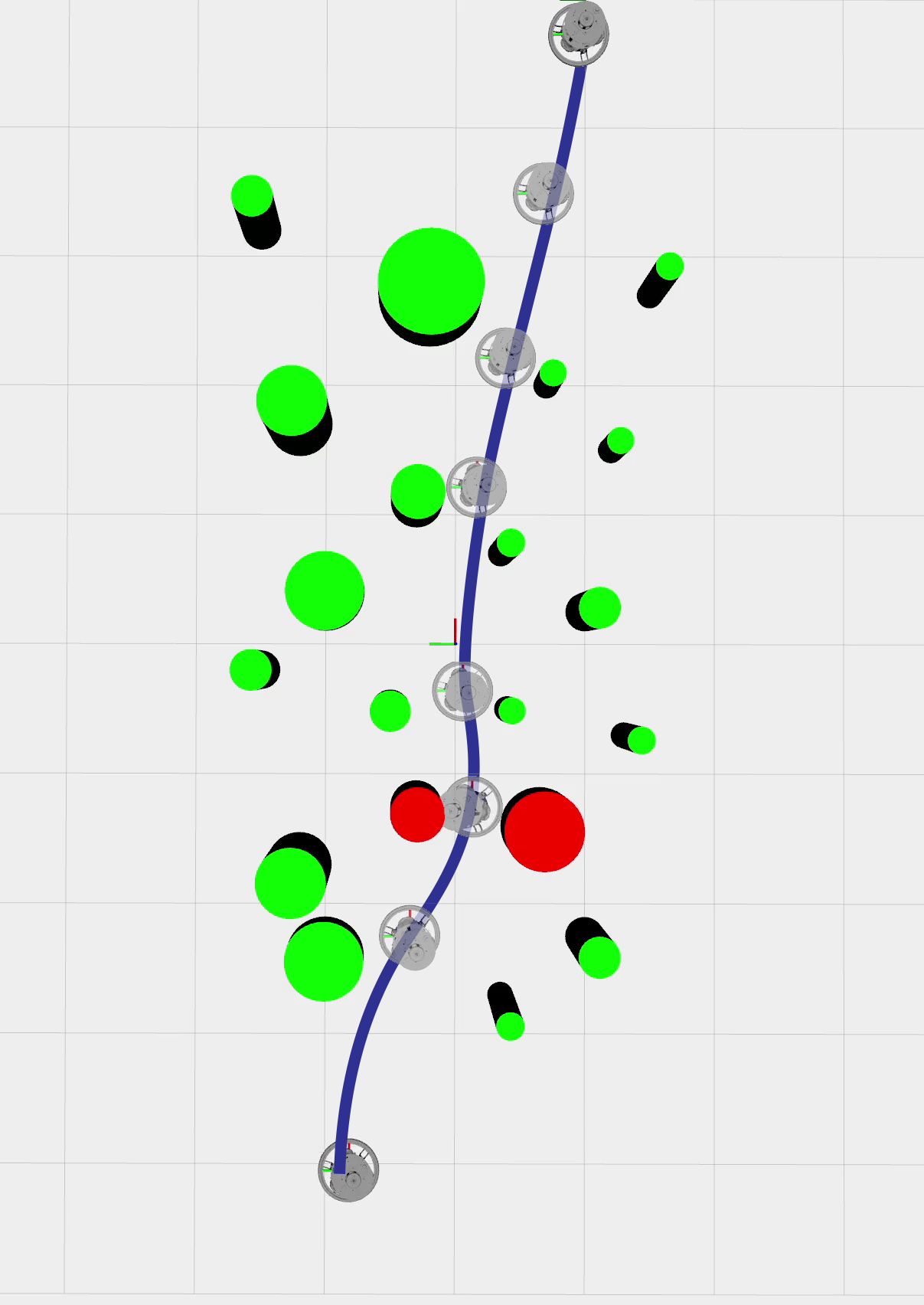}
        \caption{Collision-free ballbot trajectory computed by the three AL-methods}
  \label{fig:Obstacles}
\end{minipage}\hfill
\begin{minipage}[t]{.25\textwidth}
  \centering
\includegraphics[valign=b,scale=0.394]{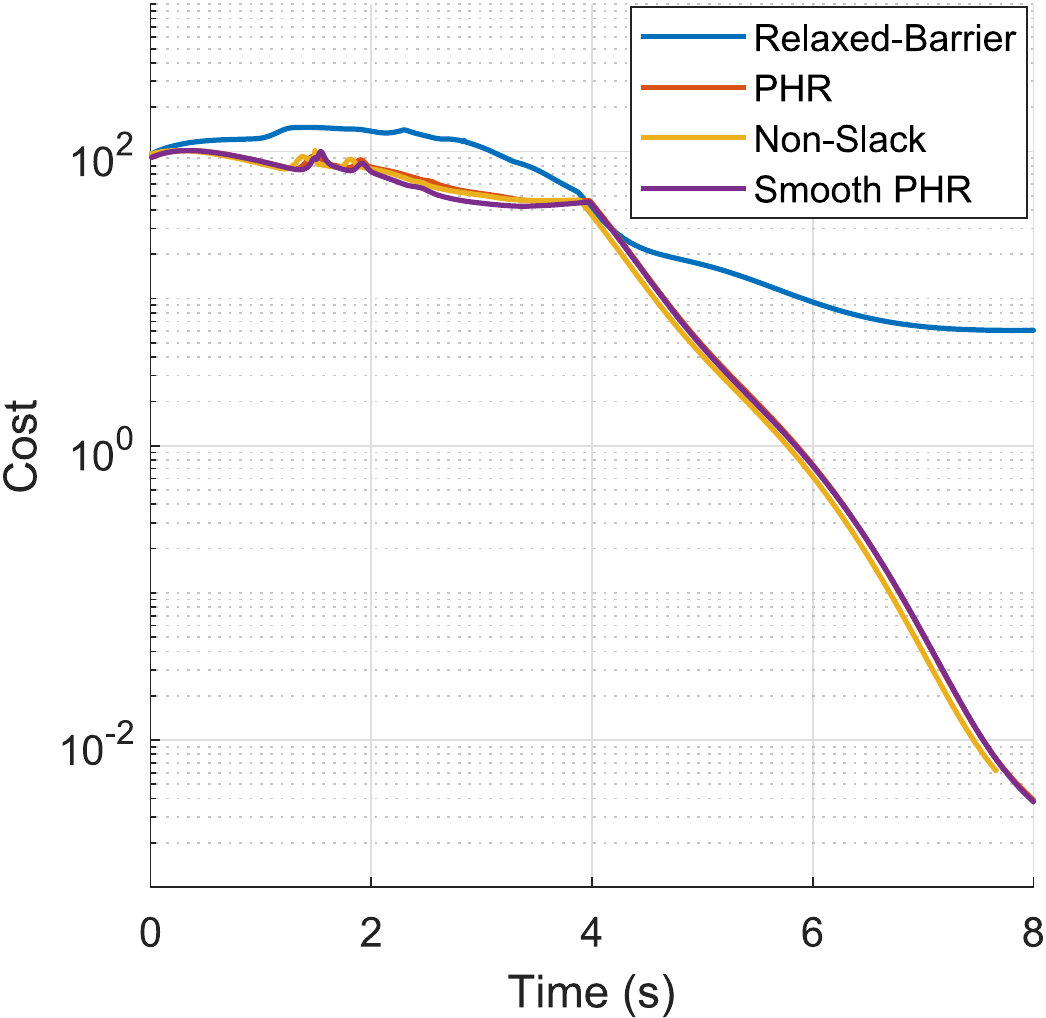}
        \caption{Plots showing the cost-function value for the four methods during the obstacle-avoidance task}
  \label{fig:Cost}
\end{minipage}
\vspace{-6mm}
\end{figure}

In this example, we impose obstacle-avoidance constraints, which are state-only non-convex inequalities, that would allow a ballbot to reach a desired goal after traversing a path surrounded by an array of 20 pillars. The ballbot system consists of 5 unactuated degrees of freedom (corresponding to the base's planar position and 3D orientation), and 3 actuated degrees of freedom responsible for actuating the supporting ball. The MPC loop runs at ${100 \ \text{Hz}}$ for a time horizon of ${T = 1 \ \text{s}}$. The AL method manages to find a similar solution for the three penalty functions with the maze presented in~\cref{fig:Obstacles}. This can be seen in their corresponding plots for the evolution of the cost function in~\cref{fig:Cost}. As for the relaxed-barrier method, the solver fails to converge when using a stiff barrier (low $\mu$ and $\delta$), and violates the constraints for an overly relaxed barrier (high $\delta$). Indeed the solver converges only for high values of $\mu$ with a low value for $\delta$. However, two consequences arise from such a choice: First, the robot is not able to pass in between the red pillars, so it gets stuck there and fails to reach the goal. This is because a high barrier parameter tightens the actual feasible region, thereby effectively reducing the gap between the red obstacles. After increasing this gap to allow the ballbot to pass through, we notice that it approaches the goal but never attains a zero steady-state error. This is because the minimizer of the new cost function is shifted with respect to the original one, thus leading to a new reference equilibrium point. The plots in~\cref{fig:Cost} further support this idea, as the original cost in the relaxed-barrier case never goes to zero. The figure also highlights the ability of AL-based methods to circumvent such an issue, thereby enhancing control performance in terms of tracking error. 

\subsection{Quadrupedal Mobile Manipulator} 
In this last scenario, we test our augmented-Lagrangian SLQ-MPC method on a high-dimensional hybrid dynamical system: a quadrupedal platform equipped with a 4-DoF robotic arm. The task description involves the robot pushing a 10 kg block from one point to another, while respecting the joint torque limits of the arm. Briefly, we use the centroidal-dynamics formulation to describe the robot's motion, and we augment that to the object's dynamics to define the full system flow map in the OCP. Therefore, the corresponding state vector is given by: ${\bm x = (\bm h_{com}, \ \bm q_b, \ \bm q_j, \bm x_o) \in \mathbb{R}^{32}}$ which includes the centroidal momentum $\bm h_{com}$, the base pose $\bm q_b$, the joint positions $\bm q_j$ and the object state $\bm x_o$; while the input vector is given by the forces acting on the limb contacts and the joint velocities: $\bm u = (\bm f_{c_1}, \ ..., \ \bm f_{c_{n_c}}, \ \bm v_j) \in \mathbb{R}^{31}$. We use a quadratic cost function where we encode the pushing task by setting a high weight on the object's deviation from the desired position. Moreover, we introduce a set of locomotion and manipulation-related state-input equality constraints that are defined at the level of the different contact points \cite{RAL}. As for the arm torque limits, those are specified as follows: ${-\bm \tau_{max} \leq \bm J^T_{c_a} \bm f_{c_a} \leq \bm \tau_{max}}$, where $\bm J_{c_a}$ and $\bm f_{c_a}$ are the arm contact Jacobian and end-effector contact forces, respectively. We choose a torque limit of ${\tau_{max} = 15 \ \text{Nm}}$ (the torques reach a maximum of ${35 \ \text{Nm}}$ during the unconstrained manipulation task). Moreover, we make the task more challenging by applying a constant fictitious external force on the block, which causes the system to start close to the boundary of the feasible region. As a result, the relaxed-barrier method either fails to converge or violates the constraints. Therefore, we exclude it from the analysis in this section. The MPC loop runs at ${60 \ \text{Hz}}$ for a time horizon of ${T = 1 \ \text{s}}$. As illustrated in~\cref{fig:AlmaBox}, and more clearly in the supplementary \href{https://www.youtube.com/watch?v=TXKNEaFvLsk}{\textcolor{blue}{video}}, the solver discovers whole-body motions that tend to drive the arm close to a kinematic singularity while pushing the block. This provides the necessary pushing force without violating the torque limits. Furthermore, we note from this experiment that for the same penalty parameter value, adopting smaller step-sizes $(\alpha_{low})$ in the outer loop generally yields better results in terms of feasibility and computational time for all three cases, as reported in~\cref{table}. This is most likely related to the Riccati solver (i.e., the adaptive step-size integrator) struggling to find solutions that meet the specified accuracy tolerances in the presence of large variations in the backward dynamics. Such variations are induced by high $\bm \nu$ values within infeasible parts of the trajectory, and low values elsewhere.        
Indeed for larger multiplier steps ${\alpha > \alpha_{max}}$, the Riccati solver fails to converge to any solution within the assigned number of allowable function calls. 
\begin{table}[]
\centering
\captionsetup{justification=centering}
\caption{\scshape results for al-slq solver during object-pushing task. the best results are highlighted in blue.}
\resizebox{0.95\columnwidth}{!}{%
\begin{tabular}{l|cccc}
\begin{tabular}[c]{@{}c@{}}\textbf{Augmented-Lagrangian} \\ \textbf{ Penalty}\end{tabular} &
\begin{tabular}[c]{@{}c@{}}\textbf{Solver Average} \\ \textbf{ Time (ms)}\end{tabular} &
\begin{tabular}[c]{@{}c@{}}\textbf{Solver Peak} \\ \textbf{ Time (ms)}\end{tabular} & \begin{tabular}[c]{@{}c@{}}\textbf{Constraint} \\ \textbf{ Violation}\end{tabular} & \begin{tabular}[c]{@{}c@{}}\textbf{Task} \\ \textbf{ Duration (s)} \\ \end{tabular} \\ \hline
\rule{0pt}{3ex}
 PHR ($\alpha_\text{high}$ / $\alpha_\text{low}$) & \normalsize	\textbf{35} / \color{blue} \textbf{25} & \normalsize	 \textbf{80} / \color{blue}\textbf{59}  & \normalsize	 \textbf{1.3} / \color{blue} \textbf{0.12}    & \normalsize	 \textbf{1.9} / \color{blue}\textbf{1.8}   \\
\rule{0pt}{3ex}
Non-Slack ($\alpha_\text{high}$ / $\alpha_\text{low}$) &  \normalsize	 \textbf{30} / \textbf{25} & \normalsize	 \textbf{65} / \textbf{55}  & \normalsize	 \textbf{1.8} / \textbf{0.70}    &  \normalsize	 \textbf{1.9} / \textbf{1.9}      \\
\rule{0pt}{3ex}
Smooth-PHR ($\alpha_\text{high}$ / $\alpha_\text{low}$)  & \normalsize	 \textbf{25} / \color{blue}\textbf{22}  & \normalsize	 \textbf{60} / \color{blue}\textbf{51}   & \normalsize	 \textbf{0.8} / \color{blue}\textbf{0.15}     & \normalsize	 \textbf{1.9} / \color{blue}\textbf{1.8}      \\
\end{tabular}%
}
\label{table}
\vspace{-3mm}
\end{table}

\begin{figure}[t]
    \centering
                \scalebox{0.65}{
       \begin{tikzpicture}
\node (A) at (0,0) {\includegraphics[trim={33cm 12cm 10cm 9cm},clip,scale=0.075,keepaspectratio]{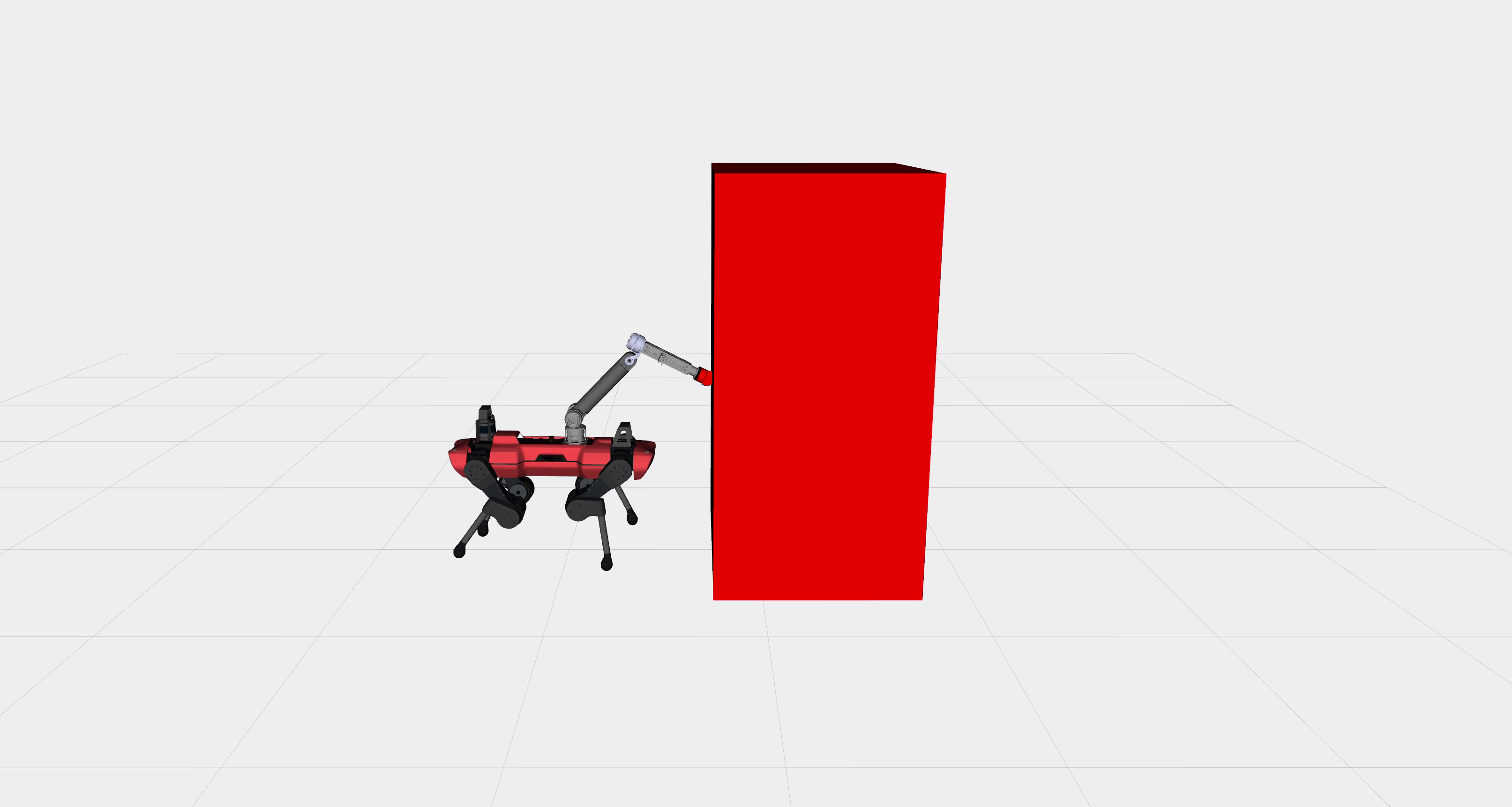}};
\node[right] (B) at (A.east) {\includegraphics[trim={33cm 12cm 10cm 9cm},clip,scale=0.075, keepaspectratio]{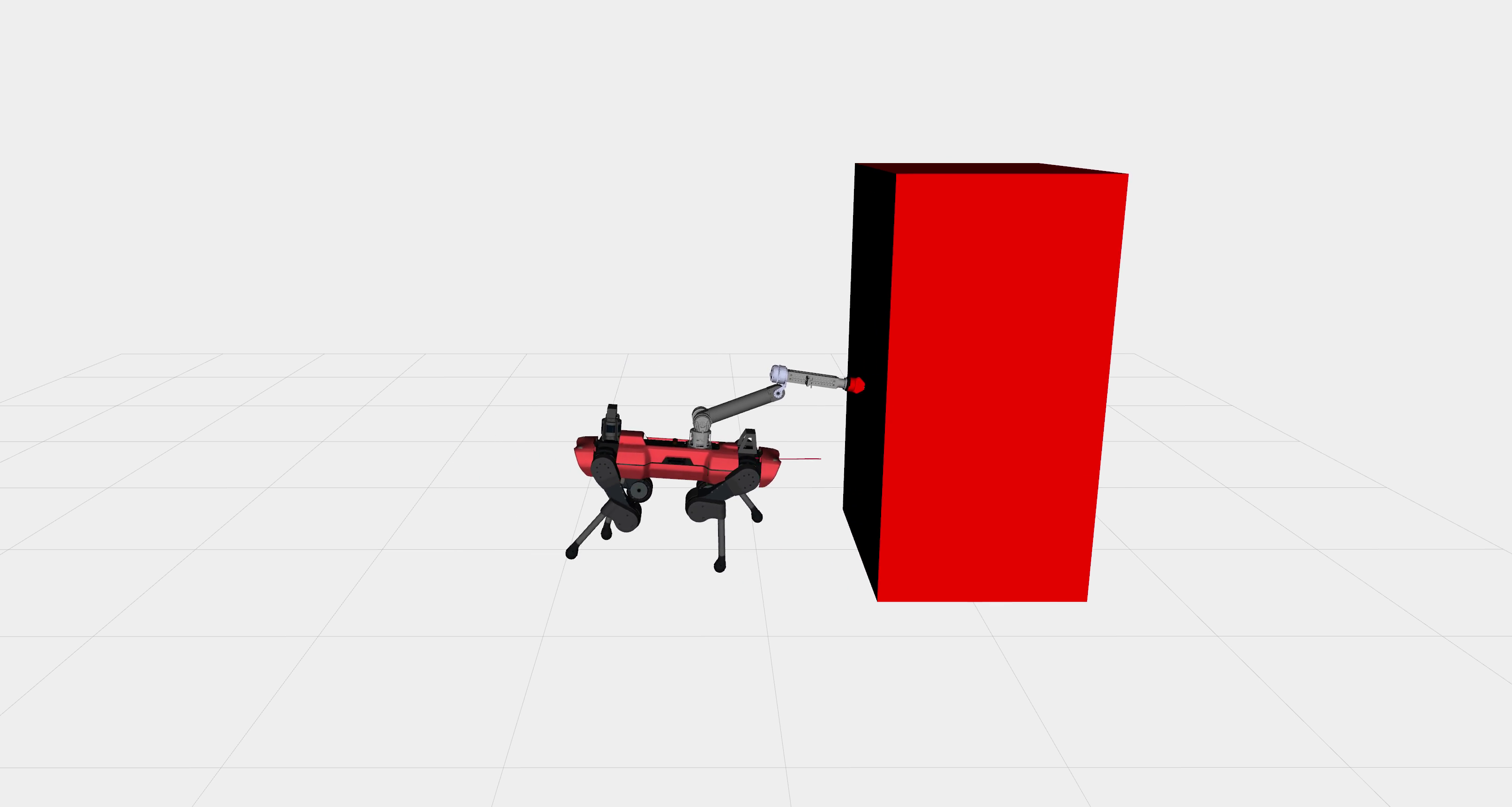}};
\node(AA) at (A.north west) {};
\node [fill=white, draw = black, minimum height = 0.3 cm, minimum width = 0.4cm, xshift = 0.675cm,yshift = -0.25cm, anchor = north east] at (AA) {\small 1};
\node(AA) at (B.north west) {};
\node [fill=white,  draw = black, minimum height = 0.3 cm, minimum width = 0.4cm, xshift = 0.675cm,yshift = -0.25cm, anchor = north east] at (AA) {\small 2};
\end{tikzpicture}}
    \caption{Snapshots of the quadrupedal mobile manipulator pushing a 10 kg load, with torque limits applied to the arm joints.}
    \label{fig:AlmaBox}
    \vspace{-6mm}
\end{figure}
\section{CONCLUSION}
We have proposed a generic constrained SLQ formulation in continuous-time, where equality constraints are handled with projection techniques, while an augmented-Lagrangian approach is introduced to treat inequality constraints. By alternating between a single inner loop and outer loop iteration, we are able to retrieve sub-optimal solutions fast enough to apply the algorithm within a real-time MPC scheme. Moreover, by interpreting the updates of the multiplier estimates as gradient ascent steps, we have been able to motivate and validate the importance of taking small step-lengths to avoid numerical issues in the solver's backward pass, thus improving convergence and stability of the solver. Three variants of AL-penalties were proposed and benchmarked in several representative robotic tasks. We conclude that the classical and smooth PHR penalties yield the best results overall. In terms of computational time, the latter has a slight advantage due to its second-order continuity. Natural extensions to this work would involve exploring alternative multiplier update methods that would potentially overcome the need for proper tuning of the step-length parameter.

\addtolength{\textheight}{0cm}   




\bibliographystyle{IEEEtran}
\bibliography{bibliography}

\end{document}